%% file: main.tex
\documentclass{article}

\usepackage{neurips_2024}

\usepackage[utf8]{inputenc} 
\usepackage[T1]{fontenc}    
\usepackage{hyperref}       
\usepackage{url}            
\usepackage{booktabs}       
\usepackage{amsfonts}       
\usepackage{nicefrac}       
\usepackage{microtype}      
\usepackage{xcolor}         

\usepackage[pdftex]{graphicx}
\usepackage{algorithm}
\usepackage{algpseudocode}

\title{Towards Vision Mixture of Experts for Wildlife Monitoring on the Edge}

\author{%
  Emmanuel Azuh Mensah \\
  \texttt{emazuh@cs.washington.edu} \\
  \And
  Anderson Lee \\
  \texttt{lee0618@cs.washington.edu} \\
  \And
  Haoran Zhang \\
  \texttt{hzhang33@uw.edu} \\
  \And
  Yitong Shan \\
  \texttt{yitonsh@uw.edu} \\
  \And
  Kurtis Heimerl \\
  \texttt{kheimerl@cs.washington.edu } \\
}

\begin{document}

\maketitle

\begin{abstract}
  The explosion of IoT sensors in industrial, consumer and remote sensing use cases has come with unprecedented demand for computing infrastructure to transmit and to analyze petabytes of data. Concurrently, the world is slowly shifting its focus towards more sustainable computing. For these reasons, there has been a recent effort to reduce the footprint of related computing infrastructure, especially by deep learning algorithms, for advanced insight generation. The `TinyML' community is actively proposing methods to save communication bandwidth and excessive cloud storage costs while reducing algorithm inference latency and promoting data privacy. Such proposed approaches should ideally process multiple types of data, including time series, audio, satellite images, and video, near the network edge as multiple data streams has been shown to improve the discriminative ability of learning algorithms, especially for generating fine grained results. Incidentally, there has been recent work on data driven conditional computation of subnetworks that has shown real progress in using a single model to share parameters among very different types of inputs such as images and text, reducing the computation requirement of multi-tower multimodal networks. Inspired by such line of work, we explore similar per patch conditional computation for the first time for mobile vision transformers (vision only case), that will eventually be used for single-tower multimodal edge models. We evaluate the model on Cornell Sap Sucker Woods 60, a fine grained bird species discrimination dataset. Our initial experiments uses $4X$ fewer parameters compared to MobileViTV2-1.0 with a $1$\% accuracy drop on the iNaturalist '21 birds test data provided as part of the SSW60 dataset.
\end{abstract}

\input{sections/introduction}

\input{sections/related_work}
\input{sections/design}
\input{sections/evaluation}

\input{./sections/conclusion}


\bibliographystyle{icml2024.bst}

\end{document}

%% file: sections/introduction.tex
\section{Introduction}
\label{sec:introduction}

\citet{tuia2022perspectives} recently presented a call for capacity building from the (tiny) deep learning community to support ecological studies as insights from environmental monitoring data can help us stay up to date with species populations that are intricately tied to our food safety, disease transport patterns, and biodiversity sustenance \citet{mora2022over, owino2022impact}. Current remote sensing technologies, however, generate massive amounts of data and often struggle in limited connectivity situations. For instance, drone flights generate hundreds of Terabytes of footage in one flight \citet{mechan2023unmanned, kong2022edge, rs14030521}. In fact, in some cases, it is more practical to fly with collected data across continents on hard drives in order to process it in the cloud \citet{youtuTinyMLTalks}, making it nearly impossible for analysis in near real time applications \citet{kay2022caltech, winkler2023activity}. 

In spite of the documented growing need for processing multimodal data on the network edge \citet{speaker2022global} with state of the art deep learning algorithms, current methods aren't reliable and energy efficient enough for the expected workload. Recent years have seen an increase in energy efficient models for machine perception, led by efficient convolution networks  \citet{sandler2018mobilenetv2, hu2018squeeze, lin2020mcunet}, with insights from Vision Transformers quickly becoming more competitive in this domain as well \citet{chen2022mobile, mehta2021mobilevit}. Given that deep learning algorithms tend to perform better with more parameters and with more varied data, advances in efficient models either use compact representations of feature space (eg spiking networks) \citet{cordone2022object}, clever ways to encode similar feature spaces of larger models (e.g. learned sparsity, low bit representation) \citet{cai2022enable, nayak2019bit}
and more recently, multimodal data \citet{xu2022mmbench}. 
Due to the inherently multimodal and time aligned nature of the data generated on the edge (drone audio-visual footage, infrared sensors, passive acoustic recorders, etc.) we believe more research for multimodal edge models without incurring excessive extra computation cost is of particular importance and that mixture of experts is a promising candidate solution.

\subsection{Choice of problem domain}

We choose ecological monitoring as our problem space because it not only provides an interesting study ground for our investigations for edge deep learning but also addresses pressing societal issues.


\textbf{Species Classification.} Species classification is well represented in the fine grain visual categorization research area due to the inherent difficulty in discriminating between closely related species, sometimes even for experts. This is evident in the long list of datasets such as \citet{van2018inaturalist, khosla2011novel, kumar2012leafsnap, wah2011caltech}. The challenge to discriminate such species near the monitoring site is further exacerbated by susceptibility to common sensor noise, and factors such as weather variability might not be observed in well curated datasets. This makes ecological monitoring data very relevant to modern edge deep learning research. 

\textbf{Connected devices on the network edge.} The concept of collaborative edge deep learning has been explored in depth in the federated learning community. However, it usually assumes replicas of the same model or different small models that independently look at different local data (data parallelism) \citet{reisser2021federated, zec2020specialized, parsaeefard2021robust, guo2020pfl}. In our work, we seek to explore how subnetworks of “a single” conventional deep model can be optimized for easy adaptation in a multi-device setting (much like training a billion/trillion parameter model using multiple GPU machines in cloud clusters because the model cannot fit on one device). This fully network-edge model parallelism approach is relatively recent, with current proposals mostly focusing on reinforcement learning approaches of splitting the model \citet{sen2023data, sen2022distributed}, while we focus on learning semantically meaningful subnetworks that can be split across devices.

\subsection{Contributions}

Our work is heavily inspired by Language Image Mixture of Experts \citet{mustafa2022multimodal}, a single tower multimodal architecture with many attractive benefits for the edge use case including conditional execution of structured subnetworks based on input data, easy dropping of redundant tokens, and co-learning of multimodal embeddings, that fits the “super model” concept we are interested in. To the best of our knowledge, this patch level conditional transformer architecture for tiny deep models has not been explored yet, with the exception of a very recent work \citet{daxberger2023mobile} which used image level mixture of experts. Our main contributions are (1) a modified model architecture of MobileViT \citet{mehta2021mobilevit} to support patch level mixture of expert computation towards per-patch single tower models (2) an evaluation of the proposed model on the iNaturalist '21 birds data in the Cornell Sapsucker Woods 60 dataset \citet{van2022exploring} for fine-grained classification of highly confused bird species. As in \citet{tuia2022perspectives}, we hope this work inspires more researchers and enthusiasts in the efficient machine learning community to participate in the high resolution ecological monitoring space towards a sustainable future for species on the planet.


\begin{figure}
  \centering
  \includegraphics[scale=0.5]{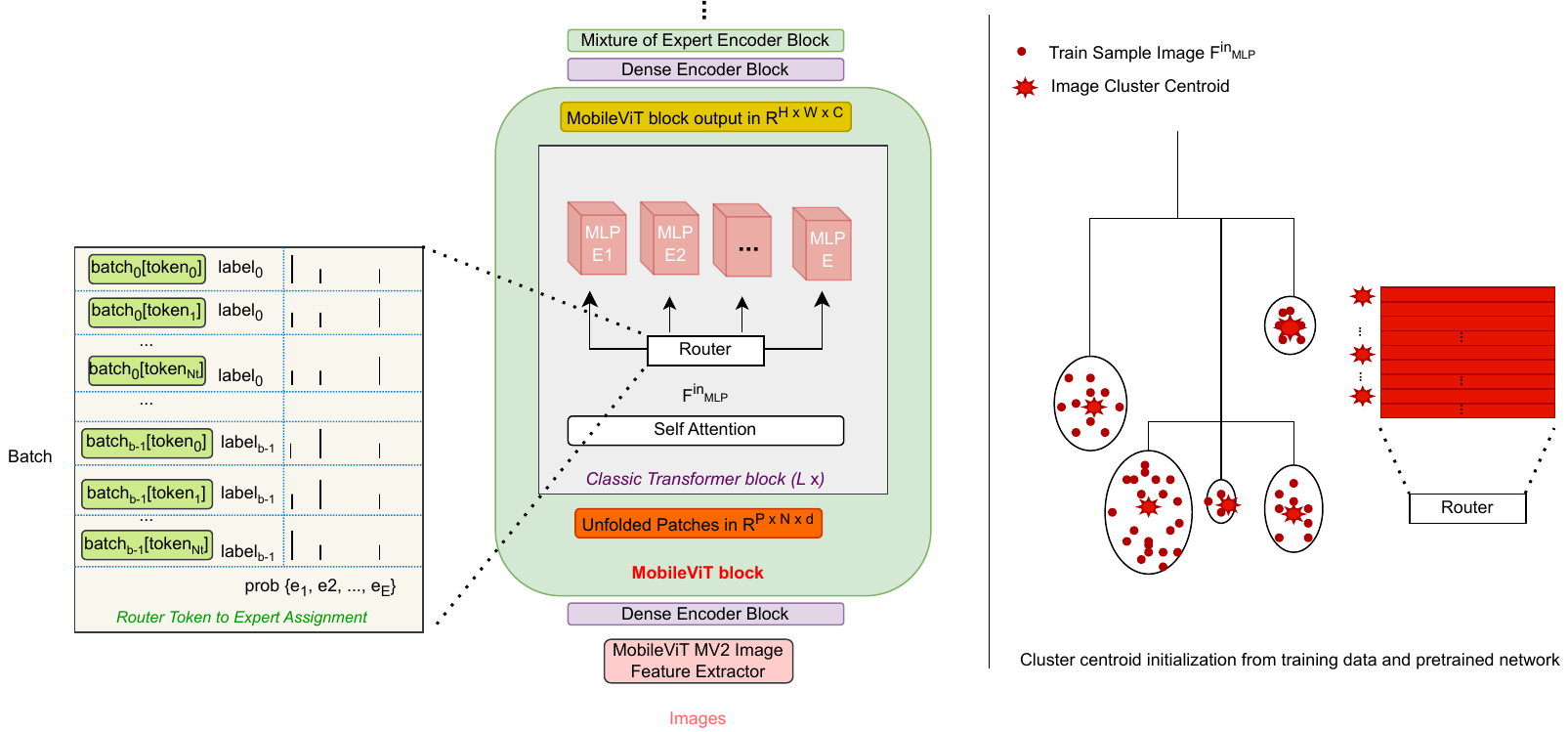}
  \caption{Our proposed system based on the MobileViT \citet{mehta2021mobilevit} model. The expert assignment router for any transformer mixture of expert layer is initialized with hierarchical clustering of sample training data embeddings collected after the attention computation of a pretrained network.}
  \label{fig:system_architecture}
\end{figure}

%% file: sections/related_work.tex
\section{Related work}
\label{sec:related_work}

Deep learning research has seen massive strides in the past decade, with a lot of established algorithms for machine perception. These foundational methods have more recently seen adoption in the tinyML community, many of which inform our work, especially when addressing the task of highly confused classification under resource constraints.

\subsection{Fine-grained visual categorization}
Fine grained visual classification has been an important part of computer vision due to its application in commerce \citet{wei2019rpc}, domestic and biodiversity monitoring \citet{van2018inaturalist, cardinale2012biodiversity}, etc. In this task, closely related categories can be very difficult to differentiate due to very large overlap in visual features between classes, sometimes needing non-visual signals as ground truth such as gene sequencing information \citet{yu2021benchmark}. For this reason, modern deep learning networks have yet to saturate the performance on existing benchmarks (e.g. $\sim{84\%}$ top-1 accuracy on inaturalist 2017 using $>$ 600M parameters) \citet{he2022masked, ryali2023hiera} and even if they do, there is still a long way to go before addressing all related issues in real deployment, such as distribution shift. Some of the major algorithms that have made great progress in this domain have included masked auto-encoders \citet{he2022masked}, coarse to fine-grain multi objective classification \citet{touvron2021grafit} and adding extra information such as geographical data in \citet{chu2019geo} which reports an increase in low resource model mobilenetV2 ($\sim 3.5$M parameters) accuracy on iNaturalist 2017 from 59.6\% to 72.2\%. Along the line of added information, general multimodal learning is a promising line of work for further performance gains.


\subsection{Mixture of experts conditional sub networks}
Recent work such as \citet{mustafa2022multimodal} have demonstrated that transformer models can be trained in a modality agnostic manner using a single model. The main proposal depends on the success of sparse conditional computation mixture of expert (MoE) models in the Natural Language Processing \citet{fedus2022switch, lewis2021base, zhou2022mixture} and vision communities \citet{riquelme2021scaling, li2022sparse} as well as  the possibility of representing practically any modality as a series of embeddings in some vector space. This opens the door to exploring the benefits of these MoEs in the edge setting. These benefits include increasing model capacity without increasing inference cost, robustness to distribution shift if coupled with a replay buffer \citet{collier2020routing}, easy dropping of many less informative tokens using  batch priority routing \citet{riquelme2021scaling} and exploiting structured sparsity of experts in edge accelerators, especially because of skipping operations in a predictable manner based on semantic information. Conveniently, there has been a number of transformer inspired edge deep models that allow exploring these benefits.

\subsection{Edge vision transformers}
Edge deep models have been dominated by convolution based networks such as mobilenet \citet{sandler2018mobilenetv2}, efficientnet \citet{tan2019efficientnet}, shufflenet \citet{zhang2018shufflenet} and so on, and even at the micro controller scale \citet{lin2020mcunet}. However, the recent widespread success of large transformer models have inspired the exploration of a new paradigm of transformer models for applications on the network edge. The most effective of them combine local inductive biases of CNNs with the global attention mechanisms in transformer models to create models competitive in accuracy, with progressively comparable FLOP count as CNNs. ConViT \citet{d2021convit} allows a flexible learning of the local inductive bias using a gated attention mechanism, whereas the smaller models use a stronger assumption on locality using standard convolutions \citet{mehta2021mobilevit, chen2022mobile, maaz2022edgenext, li2022efficientformer}

We use inspiration from these related works to propose and edge visual mixture of expert transformer.

%% file: sections/design.tex
\section{Proposed system}
\label{sec:design}
In this section, we outline design goals that influenced our choice of model architecture for this use case of in-the-wild edge machine learning.

\subsection{Ideal requirements of the learning system}

\begin{enumerate}
    \item \textbf{Robustness}. In-the-wild data is much more challenging to work with than carefully curated data \citet{wei2021fine}, hence the need for robust algorithms especially for fine grained species identification in extreme conditions.
    
    \item \textbf{Compute efficiency}. The system should ideally be easy to deploy to edge compute devices using currently supported edge machine learning libraries like tensorflow lite but can also support tiny device hardware software co-design \citet{lin2022ondevice}.
    
    \item \textbf{Cooperative execution}. Models that lend themselves to continual learning on the edge in an efficient manner are highly desirable. This is especially true because local communication might be possible (eg LoRa, \citet{dertien2023mitigating}) and sensor locations might not be completely isolated. In this case, periodic sending of relevant low data rate gradient information across monitoring sites can be more energy efficient given different capacities for energy utilization on different devices. 
    
    \item \textbf{Adaptability}. A desirable goal is to be able to triangulate ambiguous classification scenarios with more information sources. For instance, augmenting black and white images at night time with infrared camera signal or audio data. A single model capable of learning complex relationships across modalities would be very beneficial.
\end{enumerate}

If we assume the use case is in a mobile setting (or comparable devices such as raspberry pi with some access to solar power), then combining the mobile vision transformer (MobileViTV2) \citet{mehta2021mobilevit} with the conditional mixture of experts formulations satisfies these conditions.

\subsection{Model details}

As shown in Figure \ref{fig:system_architecture}, we modify the MobileViTV2 model with the mixture of experts transformer block \citet{mustafa2022multimodal} as a way to increase the capacity of the model while enforcing enough sparsity to allow low inference time latency and energy utilization. 


\subsubsection{Architecture design}
Our proposed architecture replaces the transformer layers from \citet{mehta2021mobilevit} with Mixture of Experts (MoE) as introduced in \citet{mustafa2022multimodal}. Table \ref{tab:birds_inat_vision_ablation} shows sample choices of MoE layers out of a total of 9 transformer layers in the MobileViTV2 model.

The output of the mixture of $E$ experts is the standard formulation as in \citet{mustafa2022multimodal}

\[
    MoE(X) = \sum_{e=1}^{E}{g(X)_e \cdot MLP_e(x)}
\]

where $g(X)$ is the softmax over experts in the router assignment matrix for input $X$. The multi layer perceptron (MLP) utilized in the experts transformer block is characterized by 

\[
MLP_e(x) = \sigma(x\mathbf{W_1} + \mathbf{b_1})\mathbf{W_2} + \mathbf{b_2}
\]

where $\mathbf{W_1} \in {R^{d_{layer} \times d_{ff}}}$, $\mathbf{W_2} \in {R^{d_{ff} \times d_{layer}}}$, $\mathbf{b_1} \in R^{d_{ff}}$, $\mathbf{b_2} \in R^{d_{layer}}$, $d_{layer}$ is the dimension of the input embeddings in that transformer layer and $d_{ff}$ is the dimension of the hidden unit in the transformer MLP.

\textbf{Expert router initialization} As noted in contemporary work \citet{daxberger2023mobile}, training the MoE router tends to be unstable even with large dataset sizes and large models. We encounter similar challenges and found that initializing the router with pre-determined cluster centroids obtained from layer activations of sampled training data worked well as a bootstrap in practice. Of the clustering algorithms we explored for MobileViTV2 transformer layers, Agglomerative Hierarchical Clustering with Ward linkage produced the best semantic split of data categories. 
We perform min-max scaling of pre-MLP embeddings (obtained after MobileViTV2 transformer attention and layer normalization are applied) to be used for clustering. The scaler parameters are re-used to normalize pre-router activations during training and inference. Finally, we use the Cosine Similarity between the initialization centroids and the min-max normalized activations (averaged over the pixel dimension for each patch) as the routing logits. Note however that the input min-max scaling is just for computing the routing logits and not for the expert MLP input normalization. After an expert is selected for a patch, all pixels in that patch are routed to the same expert.


\textbf{Expert centroid refinement}
Using a ``per-class'' embedding average to find the cluster centroids is compute efficient. However, when generalizing to more diverse instances of the same object, this approach introduces too much bias leading to impure routing since the embedding region between nearby centroids are not properly delineated. Using patch-level embeddings will also be unreasonably computationally expensive for most clustering algorithms. We used Algorithm  \ref{alg:algo_2} to select the top K most representative patches per class to keep compute efficiency low while improving the stability of the generated router during training/fine-tuning. We use $K=128$ and $T=5$ refining steps in all experiments. The selected images per class are randomly resized between $160 \times 160$ and $320 \times 320$ pixels at intervals of 32 to derive embeddings of different scales of objects in the images.

\begin{algorithm}
\caption{Selecting the top K most representative patches for a class by attempting to find the K of $N$ class embeddings ($X_c \in R^{N \times P \times d_{layer}}$) most similar to each other in $T$ refinement steps. }\label{alg:algo_2}
\begin{algorithmic}
\Require $X_{c}$, $K$, $T$

\State $X_c$ $\gets$ $max(X_c, dim=1)$ \Comment{{\footnotesize \textit{maximum along the pixel dimension $P$ for each patch}}}
\State $C$ $\gets$ $max(X_c, dim=0)$ \Comment{{\footnotesize \textit{initial centroid with highest values across patches.}}}
\State $S$ $\gets$ $CX_c^T$ \Comment{{\footnotesize \textit{Similarity between centroid and all class patches.}}}
\State $T_I$ $\gets$ $TOPK(S, K)$ \Comment{{\footnotesize \textit{Indices of TopK patches closest to centroid.}}}

\For{$t \to T$} \Comment{{\footnotesize \textit{Iteratively select patches in agreement with topk.}}}
    \State $C$ $\gets$ $mean(X_c[T_I], dim=0)$
    \State $S$ $\gets$ $CX_c^T$
    \State $T_I$ $\gets$ $TOPK(S, K)$
\EndFor

\State $X_r$ $\gets$ $X_c[T_I]$ \Comment{{\footnotesize \textit{TopK patches closest to centroid.}}}

\Return{$X_r$}
\end{algorithmic}
\end{algorithm}

\textbf{Expert MLP initialization}
Like in \citet{zhang2021moefication}, we select the $d_e$
most important dimensions in the transformer MLP for each expert by computing $\sigma(x\mathbf{W_1} + \mathbf{b_1})$ and selecting the top K highest dimension as permutation indices. We fix $K=2$ to reduce the size of expert MLPs. In order to compensate for the lost information due to low rank factorization, we modify the expert MLP output with a correction term ($(1 - \gamma)MLP_e(x) + \gamma X_{corr}$), where $\gamma$ and  $X_{corr}$ are trainable with $\gamma$ initialized to $0.9$ and $X_{corr}$ initialized to the full MLP output for the starting expert centroid after reversing the min-max scaling. Each MLP input uses layer normalization initialized from the MobileViTV2 MLP input layer norm.

\subsubsection{Training details}

Our training uses MobileViTV2-0.5 as the backbone with the learning rate of mixture of expert layer parameters set to $0.005$, classifier parameters to $1e^{-5}$ and the rest of the model to $5e^{-5}$. We use weight decay of $1e^{-8}$ on the classifier head and $0$ for other layers. We use a batch size of $32$ for $80$ epochs on a single \textit{Nvidia Quadro RTX 6000} GPU. Training uses AdamW optimizer with betas $(0.9, 0.99)$ for all experiments. As in \citet{mehta2021mobilevit}
we only use random horizontal flipping of input images at probability 0.5, dropout percent of $0.1$ in the classifier head and MixUp alpha of $0.2$ as the data augmentation method. Hyper parameter tuning is performed using random search of selected parameters and evaluated on a validation set. Non-MoE mobile models are tuned separately to find data augmentation and regularization parameters (most notably weight decay and dropout) for best evaluation for each model.

In order to obtain a pretrained model to convert into a mixture of expert setup for the fine grained species classification task, we first pretrained a MobileViTV2-0.5 model initialized with imagenet on the iNaturalist '21 birds data by replacing the final classification head with a randomly initialized classifier with $60$ outputs for the number of bird species.


\subsection{SSW60 dataset}
 The Cornell Sapsucker Woods 60 (SSW60) \citet{van2022exploring} is a one-of-its-kind, carefully sampled audiovisual video dataset of highly related (visually and/or auditory) 60 bird species. Compared to existing datasets, there is a lot more correspondence between the present species in the two modalities with a relatively short audio localization window (median time span of 9.96s). The dataset consists of $5,400$ videos ($3.4$K training and $2K$ test) with the corresponding species vocalization `mostly' present concurrently, although other species could be singing/calling in the same clip. In addition, there is collection of $18K$ images of the same 60 species sampled from iNaturalist 2021 dataset and a $3.8K$ unpaired audio from the Macaulay Library. 

 We evaluate our initial model with a randomly held out 20\% of the $18K$ iNaturalist '21 data dataset and train on the remaining 80 \%, with our immediate next steps targetting the audio visual use case towards in-the-wild edge insight generation with free form video data.

%% file: sections/evaluation.tex
\section{Evaluation}
\label{sec:evaluation}



In this section, we present our model performance in a low model size and low dataset regime by first demonstrating our expert initialization results, then presenting the vision performance on the iNaturalist '21 birds data. We compare our results to the benchmark vision performance for iNaturalist '21 reported in \cite{van2022exploring}. In our case, we mostly use mobile models for additional reference in order to provide more context for the MobileViT mixture of experts model performance.





\subsection{Router cluster initialization results.}
\begin{figure*}[ht!]
\begin{center}
    \begin{tabular}{llll}
        \includegraphics[width=.28\linewidth]{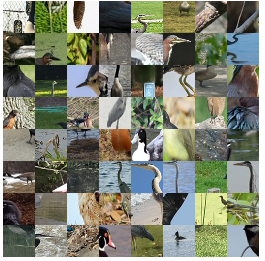} & \includegraphics[width=.28\linewidth]{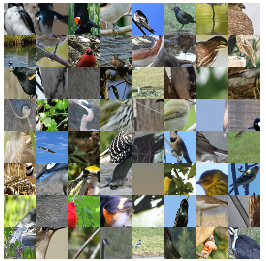} & \includegraphics[width=.28\linewidth]{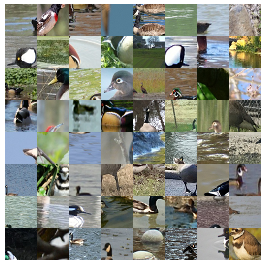}\\
        \includegraphics[width=.28\linewidth]{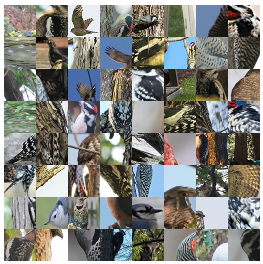} & \includegraphics[width=.28\linewidth]{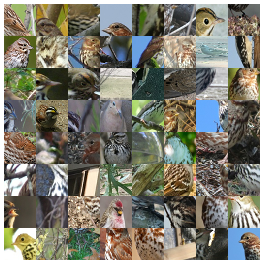} & \includegraphics[width=.28\linewidth]{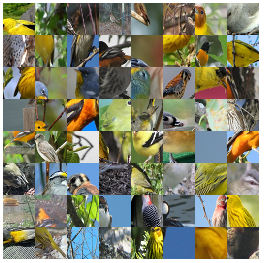}\\
    \end{tabular}
    \caption{ Sample expert groupings in the last MobileVitV2 transformer layer for iNaturalist 2021 birds species dataset included in the SSW60. It is worth noting that these are reports of relatively well behaved groupings and section \ref{sec:expert_class_affinity} discusses overall expert assignment behaviour across classes.}
    \label{fig:birds_semantic_split}
\end{center}
\end{figure*}

Since the MobileViTV2 model is trained on Imagenet, we first explored the clustering of classes with Imagenet and verified that finetuning maintained semantic splits. We then finetuned the (non-MoE Imagenet pretrained) model on iNaturalist 2021 birds species training dataset supplied as part of the SSW60 dataset. We perform similar clustering and show sample groupings of birds species at 16 expert clusters in Figure \ref{fig:birds_semantic_split}

\subsection{Inaturalist '21 birds dataset performance}

\input{./tables/birds_inat_vision_eval}

\subsubsection{Ablation experiments of MoE router parameters}

\input{./tables/ablation}

\subsubsection{Affinity between experts and class patches.}
\label{sec:expert_class_affinity}

\begin{figure*}[ht!]
\begin{center}
    \begin{tabular}{cccc}
        \includegraphics[width=.21\linewidth]{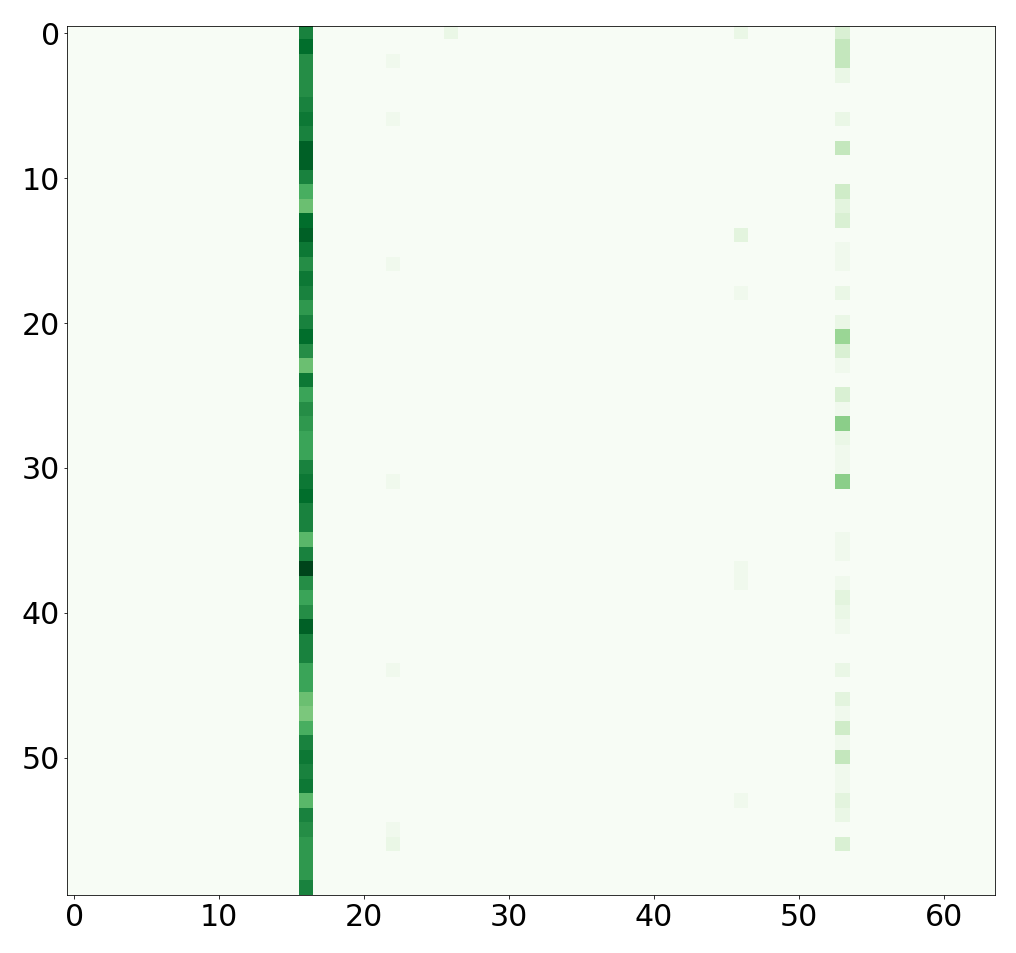} &

        \includegraphics[width=.244\linewidth]{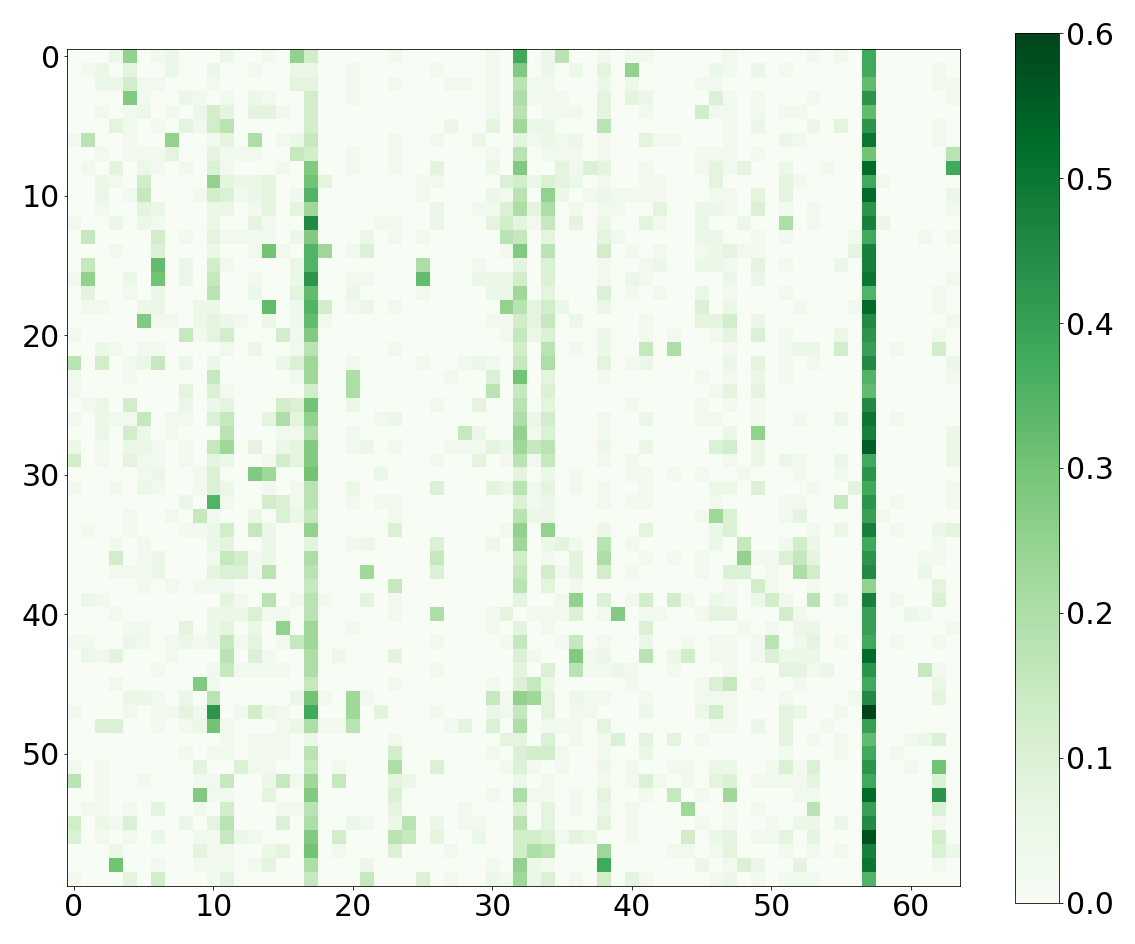} &
        
        \includegraphics[width=.21\linewidth]{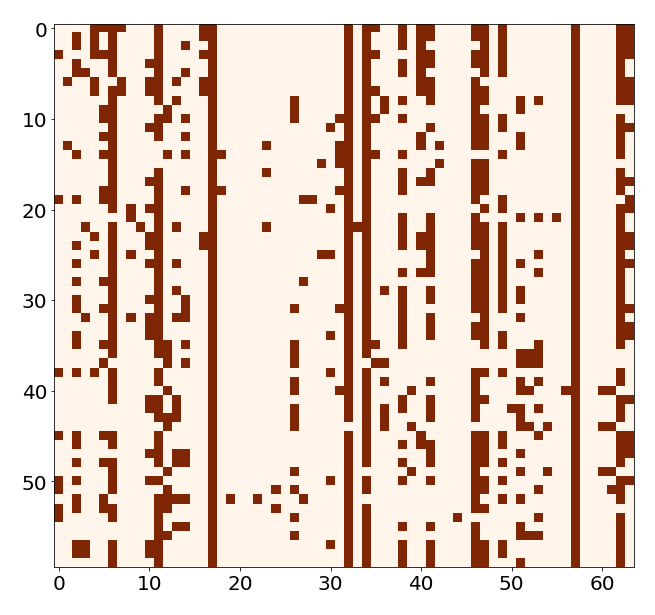} & 
        
        \includegraphics[width=.244\linewidth]{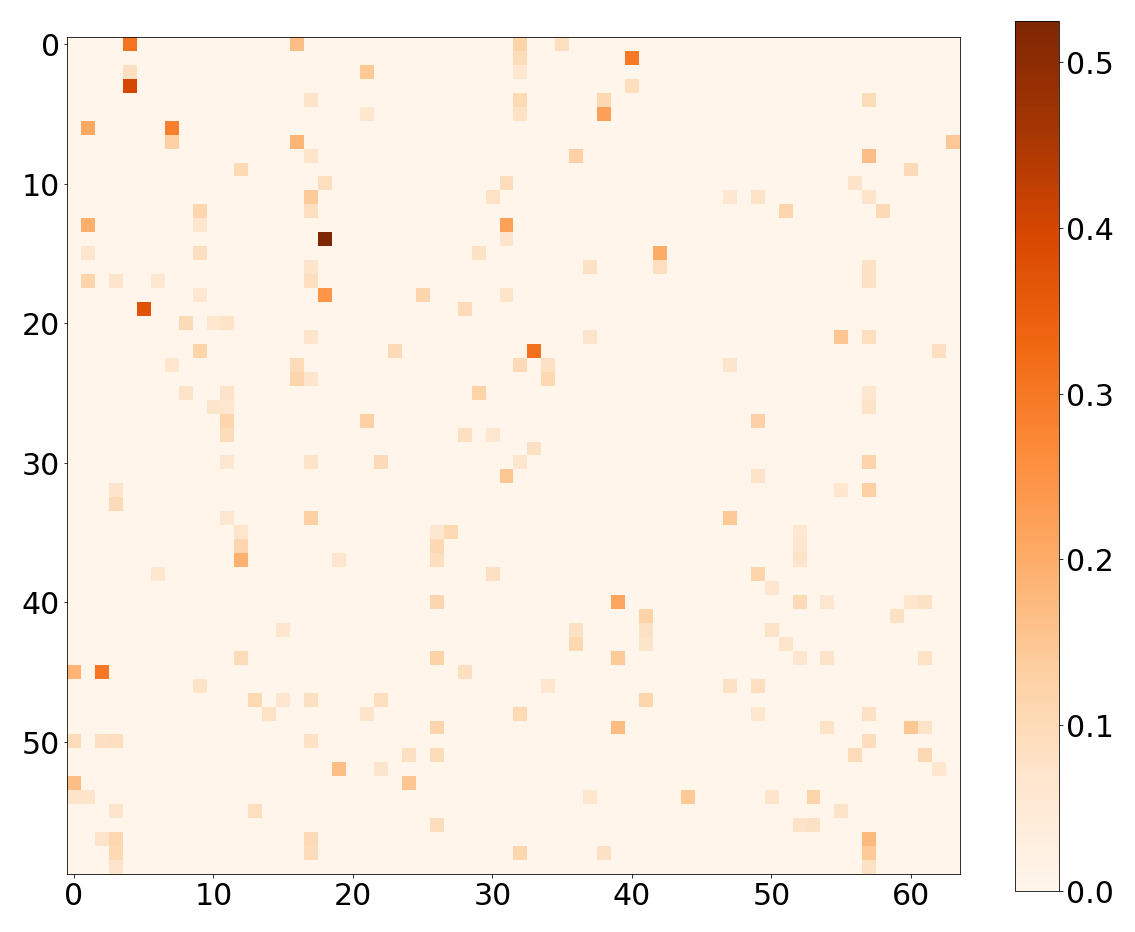}  \\
        
        $(a)$ & $(b)$ & $(c)$ & $(d)$ \\
    \end{tabular}
    \caption{ We show the expert/agglomerate cluster patch assignment probability distribution as discussed in \ref{sec:expert_class_affinity}. $a$ is the class-expert affinity after finetuning the mixture of expert model with a randomly initialized router for $80$ epochs. $b$ shows the post fine tuning class-expert affinity when the router is initialized with the cluster experts which at initialization have the affinity shown in $c$. Finally, $d$ uses a softmax temperature of $0.001$ to compute affinities and sets scores lower than $0.05$ to zero. 
    }
    \label{fig:affinity}
\end{center}
\end{figure*}

In order to better understand how the fine grained species are allocated to various experts across all classes in the dataset, we compute an affinity score for expert routing assignment after finetuning a model (green) in Figure \ref{fig:affinity} and for expert initialization clusters before fine tuning (brown) in Figure \ref{fig:affinity}. Let $G \in R^{T \times E}$ (where $T$ represents number of tokens or patches and $E$ number of experts) be derived from computing the cosine similarity between the cluster centroids and top $K$ selected patches from each class. We compute a softmax probability distribution over experts dimension and then find the per-class average for each expert. For the case of post-fine tuning affinity matrix, we collect the softmax scores for patch to expert assignments over 50 batches of randomly sampled validation data at batch size 128 and compute a similar per-class average for each expert.

From the plots in Figure \ref{fig:affinity}, we can tell that the mixture of experts model reduces the routing collapse phenomenon by utilizing the cluster initialization. However, there are still scenarios where background clusters (those with assignments from almost every class) are assigned foreground patches. This can lead to starvation of expert gradients as experts from background clusters are trained, minimizing the benefit of the mixture of expert formulation. 
In simply using the probability distribution shown in Figure \ref{fig:affinity} $(c)$ during router initialization to select which class patches should be averaged to generate the expert centroids, all class patches with allocation to a cluster contribute equally in determining the cluster centroid.

By observing that Figure \ref{fig:affinity} $(d)$ leads to more plausible affinities for foreground clusters and down-weighs background clusters in comparative terms, the cluster centroid can be refined to better capture foreground patches using a weighted average of patch contributions to computing the cluster centroid. In addition, the same affinity distribution can be used as priors when computing routing scores to re-weigh runtime evidence affinities in order to derive a bayesian posterior for expert allocation that takes into account distinction between foreground and background patches from the same class.




%% file: tables/birds_inat_vision_eval.tex
Using the expert initialization at layer 8 (the last transformer layer in MobileViTV2), we report the mixture of expert performance in Table \ref{tab:birds_inat_vision_eval}  with MobileViTV2-0.5 as the backbone model.

\begin{table}[!htbp]
    \centering
    \small
    \caption{Evaluation for training the vision models with the Inat'21 birds test dataset.}
    \begin{tabular}{cccc} \\ 
    \hline
      Model & Parameters (millions) &  Birds iNat Images \\
    \hline
     ResNet50 & 23M & 60.47\\
     \hline
     EfficientNet-B4 & 19M & \textit{70.40} \\
     MobileNetV2 & 3.5M & \textit{58.06} \\
     MobileViT-0.5 & 1.24M & 60.06 \\
     MobileViT-1.0 & 4.4M & 66.03  \\
     MobileViTMoE-2 & 1.06 & 64.70 \\
    MobileViTMoE-8 & 1.07M & 64.86  \\
    MobileViTMoE-16 & 1.08M & 65.00 \\
    \hline
    \end{tabular}
    \label{tab:birds_inat_vision_eval}
\end{table}


Although the performance scaling behavior of the current model is lower than desired, we do consistently reduce the parameter count and flop operations, while routing patches to a reliable subset of experts. Follow up work will explore incorporating class-level priors as discussed in Section \ref{sec:expert_class_affinity} to improve the foreground background distinction, which we believe can help the scaling behavior in the low data regime.


%% file: tables/ablation.tex
\begin{table}[!htbp]
    \centering
    \small
    \caption{Ablation for MoE layers using MobileVitV2-0.5 as the backbone for 16 experts at $40$ epochs.}
    \begin{tabular}{ccccc} \\ 
    \hline
      Layers & Parameters (millions) & Data Subset & Batch Size & iNat'21 Birds Images \\
    \hline
     8 & 1.088M & 1.0 & 512 & 52.16 \\
      & & 1.0 & 128 & 61.7 \\   
      & & \textbf{1.0} & \textbf{32} & \textbf{62.86} \\
      & & 0.75 & & 61.1 \\
      & & 0.5 & & 60.5 \\
     \hline
     1,5,8 & 1.065M & 1.0 & 128 & 59.9 \\
     1,3,5,7 & 1.047M & 1.0 & 128 &  60.3 \\
     \hline
     2,4,6,8 & 1.010M & 1.0 & 128 & 60.1 \\
     & & 0.75 & & 59.9 \\
     & & 0.5 & & 59.3 \\
     
    \hline 
    \end{tabular}
    \label{tab:birds_inat_vision_ablation}
\end{table}

Table \ref{tab:birds_inat_vision_ablation}, presents ablation of model properties, mainly for choice of Mixture of Experts layers and scaling parameters (batch size and data subset). From the results, we make the following observations.

\begin{enumerate}
    \item \textit{Parameter efficiency}. Since we reduce the MLPs latent dimension within each expert layer, adding more expert layers generally reduces the number of parameters. Although we generally saw best performance with experts in just the last layer, the model is able to handle successive expert layers with minimal accuracy drop, which is desirable for runtime compute efficiency in the edge use case. 
    \item \textit{Scaling power law}. According to \citet{zhai2022scaling} one needs to scale data, compute and model size in order to see improvement in performance when scaling up transformer model capacity. It is very likely that increasing the amount of training data will be needed to see consistent improvement in performance as we scale the number of experts. However, we also encourage follow up work to explore other approaches of applying non-linearity in the MLPs in order to allow few shot performance with the mixture of expert models. In particular, Figure \ref{fig:affinity} $(d)$ demonstrates further room for crafting of conditional routing. For example, simply taking the marginal likelihoods for clusters given tokens from all classes and multiplying that distribution with the gate routing probabilities could reduce the background effect and improve learning better representations even in the low data regime.
    \item \textit{Batch size effect}. Increasing batch size without a corresponding increase in the amount of data tended to reduce generalization (even with higher learning rates). This is likely in line with our observation that MLPs of popular background experts (e.g. expert 57 in Figure \ref{fig:affinity} $(b)$) end up with most of the gradient update magnitude, leading to under-utilization of increased expert capacity. In particular, the birds species dataset consists mostly of very small foreground birds, further exercabating the gradient update imbalance. A distinction of foreground vs background patches using training data statistics as in Figure \ref{fig:affinity} $(d)$ could help mitigate this issue during training by for instance dropping a fraction of background tokens with low routing confidence.
\end{enumerate}

%% file: sections/conclusion.tex
\section{Conclusion}
\label{sec:conclusion}


In this work, we present an initial form of a patch level sub computational deep learning model for application in edge deep learning. By exploring structure within contemporary vision transformer models, we propose an approach that we believe will be beneficial towards insight generation on the network edge as we move towards more sustainable computation options for multimodal sensor data. More importantly, we hope our work joins the community of proposed tiny models that not only increase access to individuals using such models but in fact enable experimentation with new forms even by using a single GPU device. Our future work will explore how to apply such mixture of expert formulation for on-the-ground species monitoring using edge devices.

\section{Limitations}
The proposed approach in its present form will need device aware optimizations in order to see actual use case deployment. For example, we use a number of transcendental operations (as seen in softmax for instance) that could benefit from some approximations. Again, the current setup of the model has many hyper parameters borrowed from the language mixture of expert community that are not explored in the paper at present. This makes it difficult to find the right parameters without a decent amount of compute resources. Labs and medium to large organizations can however afford to experiment with next directions. We also note that our current approach of deriving the mixture of expert router initialization from finetuned models might be combursome for general adoption. Finetuning a router initialized from imagenet clusters can be a decent direction to explore, although with the added implication of higher data size requirements. Finally, the current version of the model is still quite far from the `super edge' model we envision (at least in a high decoupled manner) that we hope to explore in follow up work.
